\documentclass[letterpaper, 10 pt, conference]{ieeeconf}  

\IEEEoverridecommandlockouts                              

\usepackage{graphicx, subfigure}
\usepackage{multirow}
\usepackage{xcolor}
\usepackage{array}
\newcolumntype{?}{!{\vrule width 1pt}}
\makeatletter
\newcommand{\thickhline}{%
    \noalign {\ifnum 0=`}\fi \hrule height 1pt
    \futurelet \reserved@a \@xhline
}
\newcolumntype{"}{@{\hskip\tabcolsep\vrule width 1pt\hskip\tabcolsep}}
\makeatother

\usepackage{graphicx}
\usepackage{xcolor}
\usepackage{fancyhdr}

\title{\LARGE \bf
Multi-Target Deep Learning for Algal Detection and Classification
}

\usepackage{etoolbox}
\makeatletter
\patchcmd{\@makecaption}
  {\scshape}
  {}
  {}
  {}
\makeatother

\author{Peisheng Qian$^{1}$, Ziyuan Zhao$^{1}$, Haobing Liu$^{2}$, Yingcai Wang$^{3}$, Yu Peng$^{3}$\\ Sheng Hu$^{3}$, Jing Zhang$^{3}$, Yue Deng$^{1}$, and Zeng Zeng$^{1,4}$
\thanks{$^{1}$Peisheng Qian, Ziyuan Zhao, Yue Deng and Zeng Zeng are with  Institute for Infocomm Research (I2R), Agency for Science, Technology and Research (A*STAR), Singapore}
\thanks{$^{2}$Haobing Liu is with ZWEEC Analytics Pte Ltd, Singapore}
\thanks{$^{3}$Yingcai Wang, Sheng Hu, Yu Peng, Jing Zhang are with Yangtze River Basin Ecology and Environment Monitoring and Scientific Research Center, Ministry of Ecology and Environment of P. R. China}
\thanks{$^4$ Corresponding author. The work was supported by Singapore-China NRF-NSFC Grant (Grant No. NRF2016NRF-NSFC001-111)}
}

\usepackage{cuted}
\usepackage{graphicx}
\usepackage{multicol}
\usepackage{etoolbox}
\begin{document}
\maketitle
\thispagestyle{empty}
\pagestyle{empty}



\thispagestyle{fancy}
\fancyhead{}
\lfoot{}
\lfoot{\scriptsize{Copyright 2020 IEEE. Published in the 2020 42nd Annual International Conference of the IEEE Engineering in Medicine and Biology Society (EMBC), scheduled for July 20-24, 2020 at the Montréal, Canada. Personal use of this material is permitted. However, permission to reprint/republish this material for advertising or promotional purposes or for creating new collective works for resale or redistribution to servers or lists, or to reuse any copyrighted component of this work in other works, must be obtained from the IEEE. Contact: Manager, Copyrights and Permissions / IEEE Service Center / 445 Hoes Lane / P.O. Box 1331 / Piscataway, NJ 08855-1331, USA. Telephone: + Intl. 908-562-3966.}}
\rfoot{}

\begin{abstract}

Water quality has a direct impact on industry, agriculture, and public health. Algae species are common indicators of water quality. It is because algal communities are sensitive to changes in their habitats, giving valuable knowledge on variations in water quality. However, water quality analysis requires professional inspection of algal detection and classification under microscopes, which is very time-consuming and tedious. In this paper, we propose a novel multi-target deep learning framework for algal detection and classification. Extensive experiments were carried out on a large-scale colored microscopic algal dataset. Experimental results demonstrate that the proposed method leads to the promising performance on algal detection, class identification and genus identification.



\indent \textit{Clinical relevance}—To the best of our knowledge, we are the first in the AI community to apply deep learning to detect and classify algae from large-scale colored microscopic images. The method can be easily extended and implemented for detection and classification of biological cells and tissues. 
\end{abstract}


\section{INTRODUCTION}

\begin{figure}[thb]
    \centering
    \includegraphics[width =0.45\textwidth]{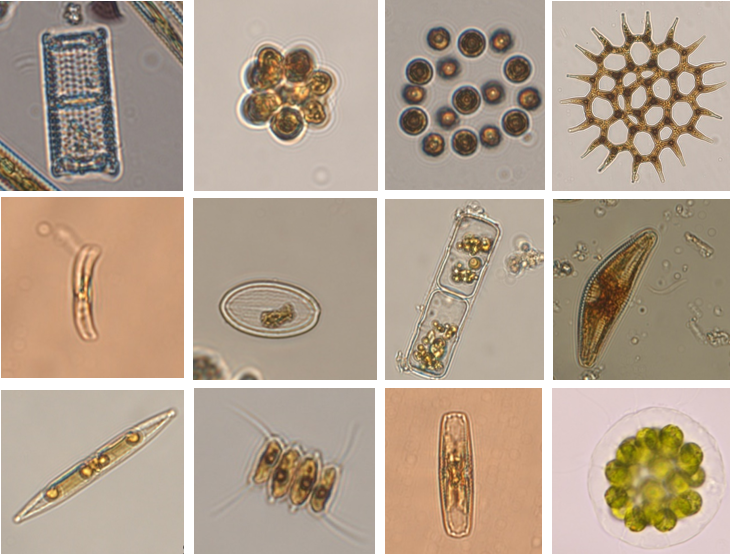}
    \caption{Examples of different algae under microscopic images. The sizes, shapes and appearances of genera are diverse.}
    \label{fig:algae_examples}
\end{figure}

Water quality is vital for modern agriculture, industry, public health and safety. Poor water quality poses risks to ecosystems and people. Biological analysis can identify changes in water quality, in which, algae are important indicators of ecological situations due to their quick responses to the qualitative and quantitative composition of species in a wide range of water conditions. Moreover, harmful algal blooms (HABs) degrade water quality and increase the risk to human health and the environment. Therefore, it is necessary to identify and enumerate algae using microscopes for water quality management and analysis. Taxonomic identification of algae is complicated, with many levels of hierarchies. For instance, an algae may belong to the genus Pediastrum, and the class Chlorophyta. Conventionally, the taxonomic classification of algae is carried out by biologists based on spectral and morphological analysis. It is a  time-consuming and error-prone task due to the micro-size and diversity of algae, see Fig.~\ref{fig:algae_examples}.

In recent years, computer vision based methods, especially the convolutional neural network (CNN), has shown promising performance in similar scenarios including object tracking~\cite{wu2019interactive}, detection~\cite{rcnn} and segmentation~\cite{zhao2019semi}. Previously, most methods~\cite{c4, c10, c3} rely on handcrafted features and limited work are presented with deep learning methods. However, these work only focused on genus-level classification, without considering the hierarchical taxonomy in biology. In this paper, an end-to-end multi-target framework extended from Faster R-CNN~\cite{rcnn, detectron2} is proposed for algal detection and classification, in which, algae positions and types are detected simultaneously, and two branches for prediction of algal genus and biological class are trained together. Furthermore, extensive experiments are conducted on a large-scale algae dataset containing 27 genera. Experimental results show the effectiveness of the proposed framework, achieving $74.64\%$ mAP@IoU=$50$ on 27 genera.


\begin{figure*}[t]
  \centering
  \includegraphics[width=\textwidth]{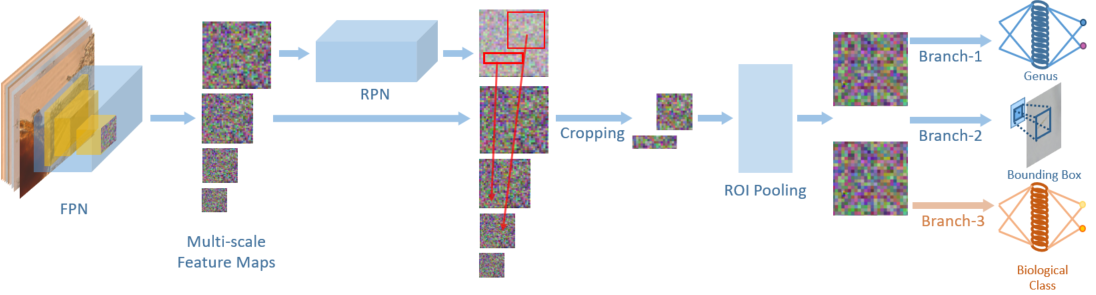}
  \caption{Model architecture. The 3 branches simultaneously outputs the genus, bounding box, and biological class of the alga. The orange component is the extended classification branch.}
  \label{fig:architecture}
\end{figure*}

\section{RELATED WORK}

Early work on the applications of machine learning in algae classification can date back from 1992~\cite{c4}, in which, a feedforward neural network was trained using OPA (Optical Plankton Analyser) features of 8 algae, achieving an average classification accuracy of over $90\%$, without graphic features. Giraldo-Zuluaga~\emph{et al.}~\cite{c2} proposed to use the segmentation algorithm to filter, orient, and subsequently extract the micro-algae profiles from microscopic images for micro-algae identification, which reported an accuracy of 98.6\% with Support Vector Machine (SVM) and $97.3\%$ with Artificial Neural Network (ANN) with 2 hidden layers, respectively. Li~\emph{et al.}~\cite{c10} employed Muller matrix (MM) imaging which highlights different biological and machine learning techniques for discrimination and classification of micro-algae. These methods highly depend on handcrafted features or extracted features from image processing, which lack generality in different environments. 

In recent years, the convolutional neural network (CNN) has been widely applied in image identification and analysis due to its ability to extract deep features from images~\cite{zhao2019bira}. For algal image analysis, only a few studies were reported using CNNs~\cite{c3,deglint2,lakshmi2018chlorella}. For example, Deglint~\emph{et al.}~\cite{c3} implemented a deep residual convolutional neural network and achieved a classification accuracy of 96\% on six algae types. An ensemble of networks was conducted on the same dataset by Deglint~\emph{et al.}, which reached 96.1\% classification accuracy~\cite{deglint2}. In~\cite{c1}, CNN models were developed by Park~\emph{et al.} from neural architecture search (NAS) for algal image analysis. In these methods, algal images were segmented from photographic morphological images with image processing methods, and classified with CNNs. These methods can not be performed in an end-to-end manner in real-world environment~\cite{cloudcomputing, ICPADS04}. Besides, the nature of algal taxonomic analysis is a hierarchical multi-label classification problem. However, in these work, algae are only classified at genus level, which does not utilize taxonomic relationships among different hierarchies. It is well noted that deep neural networks (DNNs) with the same training data samples can benefit from shared learning paths and joint learning by leveraging useful information in multiple related tasks. Such designs of deep neural networks with multiple tasks are referred to as Multi-Target DNNs (MT-DNNs)~\cite{multi-target}, in which, multiple branches are combined for stable training and better performance by exploiting commonalities and differences across tasks. 

\section{METHODOLOGY}

Inspired by MT-DNNs, The proposed framework is designed based
on the architecture of Faster R-CNN~\cite{rcnn}, as depicted in Fig.~\ref{fig:architecture}. We extend Faster R-CNN by adding an extra classification branch for multi-task learning. The framework consists of three branches that will be trained with different objectives for robust algal analysis:
\begin{enumerate}
    \item Branch-1 is used to predict the genus of algae. 
    \item Branch-2 is used for algal detection and localization.
    \item Branch-3 is used to predict the class of algae.
\end{enumerate}

Specifically, the classification of algal genus is performed in the last fully connected layer of Branch-1 based on selected regions within the bounding boxes generated from Branch-2. The last fully connected layer of Branch-3 is implemented for the classification of algal class with the shape of $m\times n$. $m$ is the number of output features, and $n$ is 6 (5 algal class, and the rest categorised as ``Others", as shown in Table \ref{table:2}). Cross-entropy is used as the loss function for algal classification at genus level and class level, termed as $L_{genus}$ and $L_{cls}$, respectively. Combined with the bounding box regression loss in algal detection, termed as $L_{box}$, the total loss function can be defined as

\begin{equation}
L_{total} = L_{box} + L_{genus} + \lambda*L_{cls} \label{eq:1}
\end{equation}

where $\lambda$ is introduced to scale the second classification loss and limit fluctuation in training. When $\lambda = 0$, the network is equivalent to Faster R-CNN.


\section{EXPERIMENTS}
\subsection{Dataset and Implementations}
The dataset used in this paper is collected from Yangtze River Basin Ecology and Environment Monitoring and Scientific Research Center, P. R. China. The dataset consists of 1859 high-resolution microscopic images of 37 genera of algae in 6 biological classes and annotations of genus and class. Some samples of algae in the dataset are shown in Fig.~\ref{fig:algae_examples}. The images were taken under microscopes with warm lighting. The color information is stored for more information compared to commonly used grayscale images~\cite{c3}. $99.2\%$ of the images have much higher resolution of $2752\times 2208$ or $3072\times 2048$ than images with resolution of $150\times 150$ used in~\cite{c1}. Furthermore, comparing to other datasets with single-alga images, our dataset is more informative and challenging. In each image, various algae of different sizes and orientations are visible, along with other noisy objects, such as bacteria, which influence the model performance. Algae were located and identified with bounding boxes. Genera and classes of algae were annotated by professionals with expert knowledge.

The dataset is highly imbalanced. Therefore algae with less than 10 instances are categorized into the existing ``else" genus, which lead to 27 genera of algae in total. The dataset is randomly split into 80\% for training purposes and 20\% for testing. The images are resized into $800 \times 800$, followed by standardization with mean and standard deviation of the dataset. The images are randomly rotated by $\pm90$ degrees and randomly cropped during data augmentation process before training.

The proposed framework is implemented using Pytorch and is trained on 4 NVIDIA Quadro P5000 GPUs with a batch size of 32, using the stochastic gradient descent (SGD) optimizer with a momentum of 0.9. The initial learning rate is 0.02, which decayed by 10 in step 6000 and 7000, respectively. ResNet-50 based FPN network pre-trained on ImageNet is applied to initialize the proposed framework~\cite{fpn, krizhevsky2012imagenet}. For algal detection, aspect ratios of the anchors (height/width) are set to $[0.25, 0.5, 1, 2, 4]$, while anchor sizes are set to $[32, 64, 128, 256, 512]$, in order to fit all sizes and shapes of algae.






\subsection{Results and Discussion}

To evaluate the performance on algal detection, mAP@IoU=$50$ (mean average precision of predictions with IoU $>=$ 50\%, abbreviated as mAP) is calculated~\cite{rcnn}, in which, IoU refers to the intersection of the predicted bounding box and the ground truth over their union. Average classification accuracy (ACA) is calculated for classification tasks. 

\begin{figure}[thb]
\includegraphics[width=0.49\textwidth]{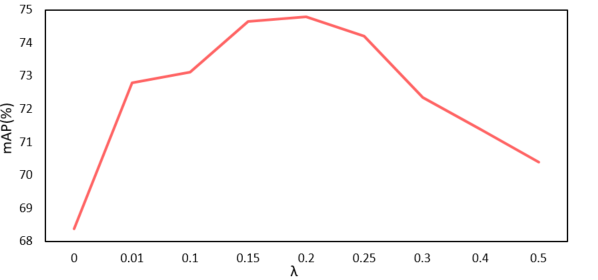}
\caption{Experimental results with different $\lambda$}
\label{fig:lambda}
\end{figure}

\begin{figure}[thb]
\includegraphics[width=0.49\textwidth]{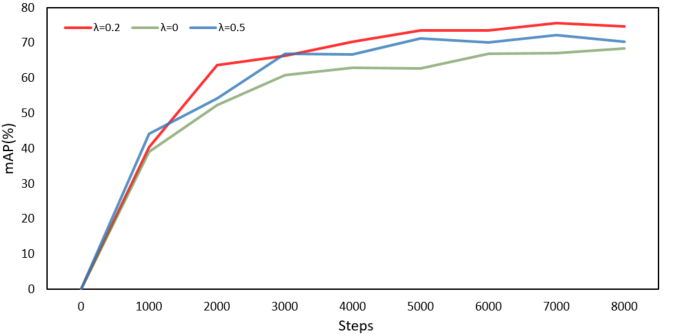}
\caption{Evaluation process with different $\lambda$ during training}
\label{fig:eval_step}
\end{figure}

The performance of the framework is affected by the value of $\lambda$ in the loss function. Therefore, we first carried out experiments with various $\lambda$ values from 0 to 0.5 to observe changes in the detection performance. The results are plotted in Fig.~\ref{fig:lambda}. It can be found that the performance is maximized when $\lambda = 0.2$. It is well noted that when $\lambda = 0$, the loss for biological class is neglected, and the third branch is not involved in the multi-target training. In this case, the model performance is worse than experiments with other settings. The comparison proves the effectiveness of our proposed multi-target learning framework. More details of the training process can be found in Fig.~\ref{fig:eval_step}. When the biological class branch is updated during training, the model consistently outperforms the baseline model (where $\lambda$ is zero) on the convergence speed as well as the final performance. Besides, we notice that when $\lambda$ value is too high, there are more fluctuations and a risk of gradient explosion during training, as presented in Fig.~\ref{fig:eval_step}. Therefore, we set $\lambda = 0.2$ in the following experiments.

The experimental results are shown in Table~\ref{table:1} and \ref{table:2}. The proposed framework achieved an average mAP of $74.64\%$ and $81.17\%$ on algal detection on the basis of genera and class, respectively. In the classification tasks, the ACA for is $96.5\%$ for genera and $99.4\%$ for biological class. Table \ref{table:1} presents the detailed detection results and the instance percentage of each genera. On one hand, 8 genera with at least $3.79\%$ of total instances in each genus, receive higher mAP than others. On the other hand, among the rest of 19 genera where each genus comprises less than $1\%$ of the data, the detection mAP is significantly low. The above mentioned results indicate that the model is sensitive to the percentage of algal instances on genus identification.

\begin{table}[thb]
    \centering
    \caption{Experimental results on algal detection at genus level. Per-genus mAP and the corresponding percentage of instances are shown. The 8 genera with highest instance percentage are presented separately. The rest 19 genera are combined into the second last row.}
    \label{table:1}
    \begin{tabular}{c?c?c}
        \textbf{Genus} & \textbf{mAP(\%)} & \textbf{Instance Percentage (\%)} \\
    \thickhline
        Cymbella & 85.89 & 28.30\\
        Navicula & 74.46 & 16.50 \\
        Synedra & 82.65 & 15.02 \\
        Achnanthes & 72.85 & 8.40\\
        Scenedesmus & 84.91 & 5.46 \\
        Pediastrum & 100.00 & 4.34 \\
        Cyclotella & 88.27 & 3.79\\
        Diatoma & 90.04 & 4.01\\
        The rest 19 genera & 22.85 & 14.20\\
    \thickhline
        \textbf{Total} & \textbf{74.64} & \textbf{100}\\
    \end{tabular}
\end{table}

\begin{table}[thb]
    \centering
    \caption{Experimental results on algal detection at class level. The mAP by biological class and corresponding instance percentage are presented.}
    \label{table:2}
    \begin{tabular}{c?c?c}
        \textbf{Biological Class} & \textbf{mAP(\%)} & \textbf{Instance Percentage(\%)} \\
    \thickhline
        Bacillariophyta & 81.29 & 82.11\\
        Chlorphyta & 86.93 & 12.78 \\
        Cyanophyta & 72.45 & 1.56 \\
        Cryptophyceae & 99.18 & 1.11\\
        Cyanobacteria & 82.14 & 1.00 \\
        Others & 25.46 & 1.44 \\
    \thickhline
        \textbf{Total} & \textbf{81.17} & \textbf{100}\\
    \end{tabular}
\end{table}

In Table~\ref{table:2}, detection at biological class level reaches $81.17\%$ mAP. It is higher than the mAP at genus level by $6.53\%$. Moreover, the detection performance is more consistent over classes with large and small instance percentages, compared to Table~\ref{table:1}. The reasons are summarized as below. First, classification at the biological class level is less fine-grained, as the biological class is a higher level than genus in taxonomy. Second, inter-genera similarity can lead to misclassification. Detection on the biological class level avoids differentiation of such similarity, and thus achieves higher mAP. Based on the classification results in Table~\ref{table:2}, we can further deduce the genus of an alga, and reach higher accuracy by exploiting the hierarchies in the biological taxonomy. ``Others" is an exception. It contains diverse classes without shared visual features, which negatively influences the classification performance.

\begin{figure}[thb]
  \centering
  \includegraphics[width=0.48\textwidth]{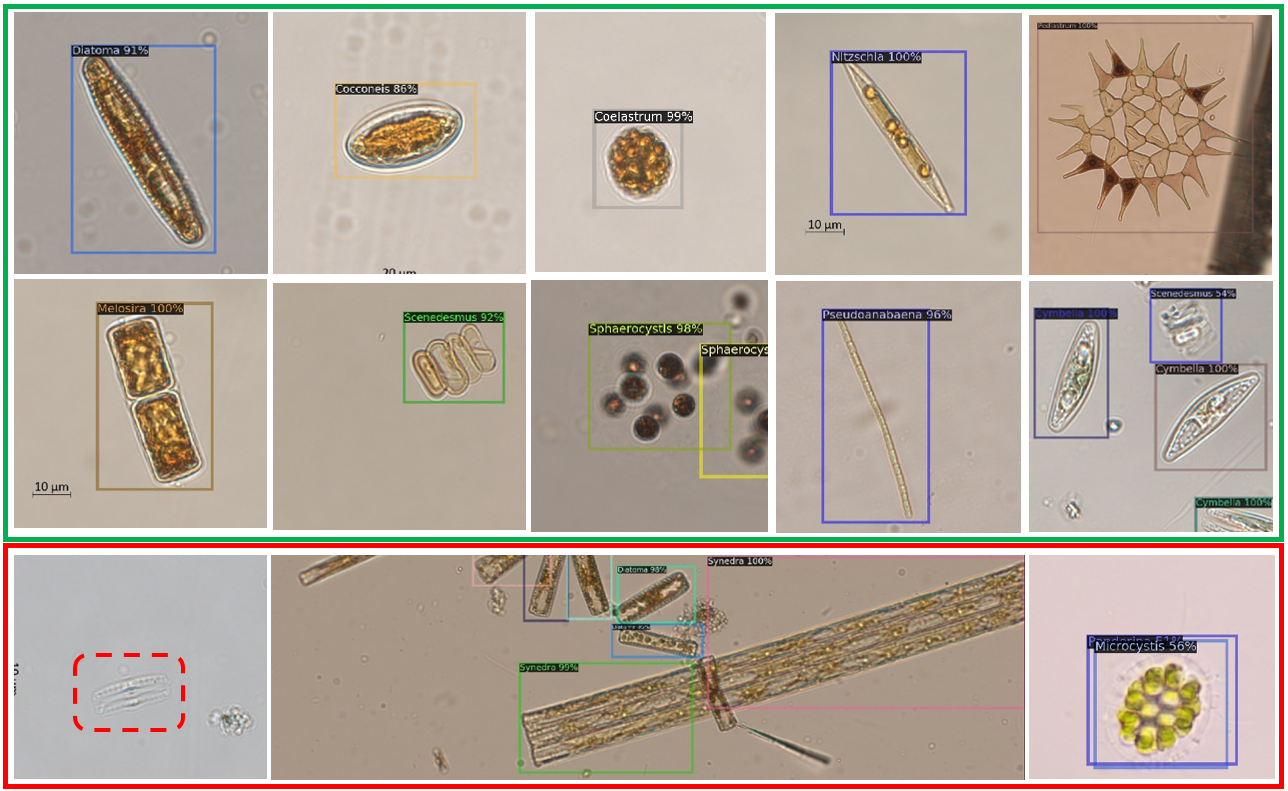}
  \caption{Examples of algal detection results. The classification results are printed on the top left corner of the bounding boxes. The first 2 rows denote correct classification and detection results. The 3 images in the bottom row represent 3 types of incorrect predictions. The error types from left to right are: undetected algae; algal occlusion and incorrect classifications.}
  \label{fig:results}
\end{figure}

Fig.~\ref{fig:results} shows some examples of good and poor detection results by our approach on the algal dataset. We find out that the proposed method is robust on most testing images, while the performance decreases under the following situations. First, undetected algae. Most of the undetected algae are almost transparent and blends into the background. Second, algal occlusion. Some algae overlap with others or with non-algae objects. Third, incorrect classifications. This is possibly due to inter-class similarity. In future work, more image pre-processing and post-processing techniques could be explored to tackle the above mentioned problems.

\section{CONCLUSIONS}
In this paper, we presented a novel multi-target deep learning framework for algal analysis. The proposed multi-target model simultaneously solves different tasks, {\textit{i}.\textit{e}.}, genus classification, algal detection, and biological class identification. The method exploits the relationships among the targets. Extensive experiments on the novel algae dataset demonstrate the robustness of our approach, achieving $74.64\%$ mAP on detection at genus level, and $81.17\%$ mAP at class level. In our feature work, 3D CNN~\cite{chen2018exploiting} can be implemented with other biological features to improve the performance on algal detection and classification.






\bibliographystyle{IEEEbib}
\bibliography{refs.bib}
\end{document}